\documentclass[preprint,5p,times,twocolumn,number]{elsarticle}

\newcommand{\ProjName}{SAIBench}
\UseRawInputEncoding

\usepackage{amssymb}
\usepackage{color}
\usepackage[usenames,dvipsnames]{xcolor}

\usepackage{booktabs}
\usepackage{url}

\usepackage{breakurl}
\usepackage[breaklinks]{hyperref}
\usepackage{makecell}
\usepackage[ruled,linesnumbered,vlined]{algorithm2e}
\journal{TBench}

\begin{document}
\begin{frontmatter}
  \title{\ProjName{}: Benchmarking AI for Science}
  \author[1,2,3]{Yatao Li}
    \ead{yatli@microsoft.com}
  \author[1,2]{Jianfeng Zhan}
    \ead{jianfengzhan@ict.ac.cn}
  \address[1]{Institute of Computing Technology Chinese Academy of Science, No.6 Kexueyuan South Road,Haidian District, 100190, Beijing, China
  }
  \address[2]{University of Chinese Academy of Sciences, No.19(A) Yuquan Road, Shijingshan District, 100049, Beijing, China
  }
  \address[3]{Microsoft Research Asia, Building 2, No. 5 Dan Ling Street, Haidian District, 100080, Beijing, China
  }
  

\begin{abstract}
Scientific research communities are embracing AI-based solutions to target tractable scientific tasks and improve research workflows. However, the development and evaluation of such solutions are scattered across multiple disciplines. We formalize the problem of scientific AI benchmarking, and propose a system called \ProjName{} in the hope of unifying the efforts and enabling low-friction on-boarding of new disciplines. The system approaches this goal with \textit{SAIL}, a domain-specific language to decouple research problems, AI models, ranking criteria, and software/hardware configuration into reusable modules. We show that this approach is flexible and can adapt to problems, AI models, and evaluation methods defined in different perspectives. The project homepage is ~\url{https://www.computercouncil.org/SAIBench}.
\end{abstract}
\end{frontmatter}

\section{Introduction}
Artificial Intelligence has seen continuous and significant advancements over the past years, with Deep Learning methods being arguably the most representative and focused on. Blessed by the ever-increasing computation power in AI accelerators and general-purpose architectures alike, new AI paradigms and models are proposed which greatly improve the scalability, flexibility, and applicability of this data-driven approach. As a result, the IT industry is welcoming AI-powered solutions, integrating them into existing data processing pipelines that will otherwise require human intervention or prohibitive computation cost. This trend is also propagating into scientific research communities, as researchers are gaining interest in leveraging state-of-the-art AI solutions to tackle equally if not more difficult tasks, hence AI for Science.

From a bird's eye view, a scientific research activity can be mechanical or creative. A mechanical research activity can be algorithmically specified, with quantized or computationally verifiable input/output. On the other hand, a creative research activity breaks out of a mechanical system, for example, by defining a new problem or introducing ideas hard to quantify. In this work, we call a computationally verifiable research task a ``tractable scientific task''.
That said, an AI for Science solution is introduced to bring improvements into the scientific research workflow and is usually targeting towards a tractable scientific task, such as:

\begin{itemize}
\item Mathematical Problem Solving --- to solve mathematically well-defined problems.
\item Pattern Matching --- to classify, identify patterns, and detect region-of-interest in high volume scientific data.
\item Prediction --- to compute future world states, given an initial snapshot of the world state and evolving rules.
\item Artifact enhancement --- to improve the quality of data acquired from imperfect observations, e.g.\ incomplete, fragmented, noisy sensor data.
\item Control --- to use actuators to drive sensor readings into desired states, despite the imperfection of both.
\item Hypothesis and Confirmation --- to propose a theory (e.g.\ equations) that conforms with the observations.
\end{itemize}

\begin{table}[t!]
  \begin{center}
  \begin{tabular}{ll}
    \toprule
    \textbf{Mathematical}      & Partial derivative equations \\
    \textbf{Problem Solving }  & General matrix multiplication \\
                      & Matrix decomposition \\
                      & Integration \\
                      & Monte Carlo methods \\
    \textbf{Pattern}           & Species Classification \\
    \textbf{Matching}          & Event Identification~\cite{albertsson_machine_2019} \\
                      & Climate Analysis~\cite{kurth_exascale_2018}\\
                      & Anomaly Detection \\
    \textbf{Prediction}        & High-Energy Particle Simulation \\
                      & Molecular Dynamics\\
                      & Fluid Dynamics\\
                      & Protein Folding \\
    \textbf{Artifact}          & Genome Sequence Alignment \\
    \textbf{Enhancement}       & Astronomy Image Enhancement \\
                      & Medical Image Enhancement \\
                      & MRI Reconstruction \\
    \textbf{Control}           & Tokamak Plasma Control~\cite{degrave_magnetic_2022} \\
                      & Sensor Triggering \\
    \textbf{Hypothesis and}    & Automatic Physics Laws Discovery \\ 
    \textbf{Confirmation}      & Symbolic Regression \\
    \bottomrule
  \end{tabular}
  \end{center}
  \caption{Examples of Tractable Scientific Tasks.}\label{tab:tractable_tasks}
\end{table}

Examples of these tasks are shown in table~\ref{tab:tractable_tasks}.
The term ``AI for Science'' is also conventionally deemed as an ensemble of vertical fields and tasks~\cite{argonne_national_laboratory_ai_nodate}. However, we argue that to fully realize the potential of AI for Science, it is not enough to cherry-pick an AI method, match it against a specific task, and heuristically compare it with existing methods.
One strength of AI methods is that they abstract away the problem details and mathematical procedures, into generic functions that transform inputs into outputs --- that is, every AI model possesses the potential to adapt to other tasks, some (for example, neural networks) even being universal approximators.
Science is vast, and AI methods are many. A single effort to evaluate a task-method pair would leave other research communities unaware, of both the potential tasks that a model is capable of processing and potential models that can be applied to a task. 
This problem is exaggerated by the fact that the AI research is moving forward fast, that by the time a specific method is picked up by a scientific computing task, or a task is adapted to an AI method, it may be already bested by then state-of-the-art. 

To help the scientific research communities as a whole systematically absorb and integrate the advancements of AI research, and to avoid repeated efforts in development and evaluation, we propose \textit{\ProjName{}}, a system that bridges scientific computing tasks and AI methods, and automatically benchmarks every sensible combination, collects performance metrics, and projects it back into rankings proper to each research community. Research groups of different backgrounds can focus on their needs while taking advantage of other benchmarking building blocks, without having to re-invent end-to-end evaluation processes.

The rest of this article is organized as follows. We first define the problem of scientific AI benchmarking in Section~\ref{sec:problem}. In Section~\ref{sec:methodology} we discuss the methodology, goal, and challenges. Section~\ref{sec:design} elaborates our system design, including the details of each component. We showcase end-to-end scenarios involving multiple modules in Section~\ref{sec:casestudy}.

\section{Problem Definition}\label{sec:problem}

Here we define the problem of scientific AI benchmarking. To begin with, we have a set of tractable scientific tasks as defined in the previous section, and an array of AI methods, each needs to be trained to solve a specific problem. To evaluate a method for such tasks, different scientific communities have different criteria.  For example, instruments in High Energy Physics generate zettabytes of data, and the training data for AI models is virtually unlimited. An AI method could thus focus on throughput, time-to-solution, sample selection, etc. Meanwhile, for Biology and Life Sciences, sometimes there are just a few hundred data points, requiring high sample efficiency, and a strong ability to generalize and extrapolate onto unseen problem configurations.

Nonetheless, the qualification of a method can be categorized as follows:
\begin{itemize}
\item \textbf{Defined by Problem Class.} For purely computational tasks such as mathematical problem solving, it is preferable to target against classes of problems to see how the method performs under each set of mathematical constraints. For example in equation solving, it is desired to study how a method behaves for both stiff and non-stiff systems, where both types contain their problem class definitions.
\item \textbf{Defined by Problem Setting.} Compared to purely mathematical problem classes, this type of problem definition usually embodies specific constraints under a class to match a physical setting. Scientific research communities have established well-respected problem settings that have been practiced and confirmed. This allows computational methods to interoperate with real-world experiments, as specific experimental settings can be virtually replicated.
\item \textbf{Defined by Problem Cases.} For some tasks we are only interested in a narrow range within the whole problem space. Most data-driven tasks fall into this category, where the typical workloads of a task are defined by collected and/or labeled data. There are also ``golden standards'' defined in research fields, which are computational methods with superior accuracy and other desirable properties, at expensive computational costs. These methods are then used to collect data for very specific problem cases so that other faster but less accurate methods can be developed and evaluated.
\end{itemize}

This categorization is not mutually exclusive though, as some tasks require more than one qualification criteria to properly define the problem. For example, a robotic control algorithm can be tested both in a simulated setting and on data points collected from real-world sensors. Nevertheless, the principle is that this categorization describes the hierarchy of problem definition -- the more the definition leans towards the former (problem classes), the more computation is required; On the other hand, the more towards the latter (problem cases), the more data. Furthermore, the problem definition serves as a specification for the AI model behavior, for both training and testing.

\begin{table}
  \begin{center}
  \begin{tabular}{ll}
  \toprule
  \textbf{Problem Class} & Boundary Value Problem \\
                & Stochastic Differential Equations \\
                & Many-body Interactions \\
                & Positive Definite Matrix Decomposition \\
  \textbf{Problem Setting} & Temperature and pressure dependence \\
                  & of alanine dipeptide \\
                  & Straight wire Magnetostatics\\
                  & Community Atmosphere Model \\
                  & (CAM5)~\cite{neale_description_nodate} Simulation\\
  \textbf{Problem Case}  & ANI-1x~\cite{smith_ani-1ccx_2020}, GDB-17~\cite{ruddigkeit_enumeration_2012} \\
                & OASIS~\cite{marcus_open_2007} \\
  \bottomrule
  \end{tabular}
  \end{center}
  \caption{Examples of Qualification Criterion.}\label{tab:qualification}
\end{table}

Examples of these qualifications are shown in table~\ref{tab:qualification}. However, AI-based methods likely require training, so the problem definition of all three types must be reduced to case-by-case training data points --- for a problem class, the problem definition should generate independent problem instances that sufficiently cover the problem space. For a problem setting, the problem definition should generate state snapshots that conform to the constraints. For data-driven problem cases, the problem definition should simply enumerate from the dataset.

Furthermore, the evaluation of a method depends on the problem definition generating tests. For each test case, the performance is represented with a cost function. For a mathematical problem instance, the cost function can be the error against ground truth solution, or error against equality constraints~\cite{weinan_deep_2017}~\cite{raissi_physics-informed_2019}. For simulation settings, the cost function can be obtained by comparing the performance metrics derived from such experiments, as shown by previous works on specific tasks~\cite{noe_machine_2020}~\cite{mardt_vampnets_2018}~\cite{jia_pushing_2020}. Lastly, for data-driven problem cases, the dataset can be split into training and test sets, and the cost function is the loss function applied to the test set.

Finally, it is crucial to realize that different benchmarking communities use the word ``performance'' to refer to different concepts. Scientific AI benchmarking concerns not only the accuracy of AI models but also the computation cost. The computation cost can be further broken down into two phases: 1) the cost for a model to reach certain accuracy, and 2) once the model is properly trained, the cost of using the model for inference tasks. For the first phase, the standard practice is to measure training time (wall clock or total CPU/GPU time) against the best/mean/worst accuracies, and for the second, the throughput/latency etc.~for completing the inference tasks. Moreover, for both phases, we can investigate the system performance with standard parallel computing benchmarking techniques~\cite{hoefler_scientific_2015} to expose different performance characteristics of a solution, for example, time-to-solution or cost-efficiency.

\section{Methodology}\label{sec:methodology}

The main goal of \textit{\ProjName{}} is to build an inclusive and interconnecting environment for all the relevant research efforts, including problem definition, AI method, training algorithm, software and hardware environment, metric definition and ranking definition, and deliver benchmarking result efficiently with given computation resources. The desiderata brought by this goal is multifold. 


We need to design the system with a modular paradigm and provide friendly programming interfaces for different modules. It should handle the impedance mismatches between different programming languages and environments while maintaining consistent standards. This is traditionally implemented with language bindings (for example, the computational chemistry package NWChem~\cite{apra_nwchem_2020} can either execute its own scripting language, or be controlled by a Python language binding) or file-based inter-process communication, which is suboptimal because different programming environments may have incompatible constructs that cannot be bound into a single process, and distributed computing modules cannot be modeled easily. 

A module should be self-descriptive so that the system can automatically discover benchmarking tasks it can participate in, so in addition to modular interfaces targeting benchmarking tasks, there should be a protocol for modules to exchange metadata and relate to each other. It is challenging to design such a protocol because it has to be generic, extensible, and yet carrying concrete meanings. 
For example, if we model the input/output of an AI model as tensors of required dimensions, it places strict constraints on what the AI model can solve, and the system will not be able to associate this AI model with even slightly incompatible tensors, not to mention non-tensor data that has to be converted to adapt. On the other hand, if we simply attach a textual description to each module, it would be too hard for machine-understanding, and require human intervention to develop the connections. To this end, machine-understandable flexibility and extensibility are needed, to enable modules to cooperate less rigidly.
The previous example shows how a module for an AI model should describe itself. Similarly, for a problem definition, it should programmatically set up the training and test fixtures, and conduct the experiments. This way all the three types of problem definitions previously can be normalized and become accessible to AI models. In addition, it should expose metadata that allows the system to inspect the execution workflow, and identify tasks that can be completed by other modules. This type of meta-programming is practiced in programming language research and recently machine learning frameworks, implemented in declarative languages and domain-specific languages (DSLs)~\cite{paszke_pytorch_2019}, yet largely unexplored in scientific computing, where most execution engines take a parse-interpret-execute approach~\cite{brehm_sanscript_nodate}\cite{apra_nwchem_2020}. 


As we discussed above, the system is not a single benchmark, but a collection of such, to be projected back to each research field and aggregated by a ranking criterion. Conflict of interests naturally arises, for example, to favor speed vs.\ to favor accuracy, first principle metrics vs.\ a particular set of derived properties. The system should be able to allow different perspectives of the same metrics and provide an interface for ranking modules to declare their preferences.

The performance of an end-to-end AI solution to a tractable scientific task depends on multiple aspects, including the AI model, the training algorithm, the computing software stack, the empowering hardware, and so on. These factors do not contribute to the final performance linearly, for example, a particular AI model may have the best work-precision properties under one hardware configuration but not the others. It is thus desirable to consider all these factors as benchmarking hyperparameters. There are several implications brought by this requirement. The AI module implementation should be declarative instead of being bound to a specific software/hardware stack; The software stack module should declare the capabilities (e.g.~matrix multiplication and backward propagation) so that the system finds compatible model-software pairs; Also, the software stack module should describe the hardware compatibility and accept a standardized hardware configuration descriptor, so that the system can automatically schedule scalability tests.

\begin{figure*}[t!]
  \centering
  \includegraphics[width=\textwidth]{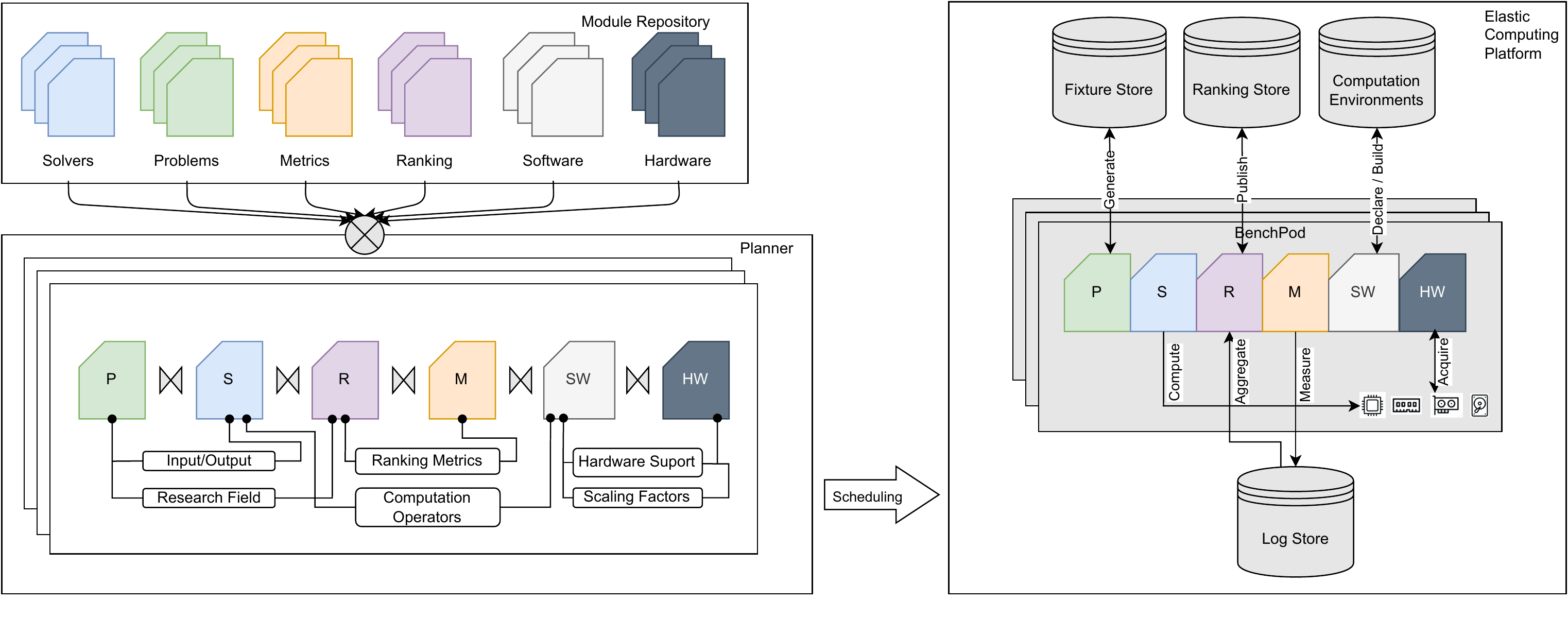}
  \caption{System Architecture.}\label{fig:design_overall}
\end{figure*}

With all the components modularized and parameterized, the whole benchmarking workflow can be formulated as follows. Each type of module introduces some \textit{dimensions} to the benchmarking task, and the goal is to enumerate and test against the Cartesian product of all such dimensions, where each point in the problem space represent the combination of a specific task, solver, metrics, software and hardware configuration. This allows the modules to advertise themselves, discover the others, and therefore reuse data and interact with each other, without knowing them beforehand. This paradigm aligns well with the FAIR guiding principles for scientific data management~\cite{wilkinson_fair_2016}, which suggests that scientific data should have findability, accessibility, interoperability, and reusability. 
This is the key difference between the methodology of this work and previous AI benchmarking and scientific benchmarking systems, where the benchmarked scenarios are pre-determined workload and model combinations, and the addition of a new AI model or dataset would not be automatically discovered and reused by existing modules in the system and has to be scripted by a programmer.

Last but not least, because the system automatically discovers potential benchmarking tasks, it is desired that the system can concurrently schedule computation resources to them. As different benchmarking tasks may require different computing environments, it is crucial that the system can elastically provision the environment for each task in a standardized manner, including the operating system, runtime libraries, setup scripts, and test fixture data. The challenge lies in how to design the system to efficiently support such needs and minimize the deployment overhead.

\section{System Design}\label{sec:design}

In this section, we illustrate the overall design of the system and tap into each system component, and discuss how to address the aforementioned challenges. The architecture of the system is depicted in figure~\ref{fig:design_overall}. The workflow is straightforward. The planner pulls all modules from the module repository and joins them into feasible tuples according to the metadata descriptors. The execution plan is then dispatched to the elastic computing platform which provides storage, processors, and accelerators, where each benchmarking task tuple is executed in a ``benchmarking pod''. The purpose of the BenchPod is to provide task-level isolation to computation resources, a communication endpoint to interact with the planner, and experiment orchestration. A problem definition module either generates data on-the-fly or retrieves a well-known dataset into the BenchPod instance. The hardware definition module acquires hardware resources. The software definition module constructs a containerized environment, based on a standardized software package requirement descriptor. The entry point of the container is a shim program provided by the BenchPod instance that orchestrates the actual execution of the solver, metric collection, and aggregation. 

\subsection{SAIL: Scientific AI domain-specific Language}
Previous AI benchmarking systems either implicitly define a series of built-in modules~\cite{zheng_aibench_2019}~\cite{mattson_mlperf_nodate}, or expose a markup language schema to define modules~\cite{chang_mlharness_2021}. For better programmability, discoverability, and user ergonomics, we propose to define modules with an embedded domain-specific language (eDSL) called SAIL. The eDSL is implemented as a Python package so that a module implementer can take advantage of modern IDE features such as auto-completion and type checking while writing the module definition. 
To design an eDSL means that the desired features must be retrofitted into the target language. To achieve this, we take advantage of various Python language constructs that best fit the required features. Some features can be implemented with static analysis, for example, we use Python decorators to identify module entry points. This way we can easily scan for modules with reflection, and build our module repository. We use Python classes to represent type descriptors for our type system, which is a dual-role construct that both encodes the type information for static analysis, and dispatches code during benchmarking runtime. Benchmarking concepts are modeled as well-known global objects, and the methods attached to them represent benchmarking primitives. This gives a hint to the user that these concepts are stateful, and the primitives can function as both computation routine and data storage. Finally, we use declarative methods to construct the computation graph for AI models. Table~\ref{tab:language} shows some language construct examples.

\begin{table}[t!]
  \begin{center}
    \begin{tabular}{p{0.11\textwidth} p{0.11\textwidth} p{0.17\textwidth}}
    \toprule
    \textbf{Feature} & \textbf{Construct} & \textbf{Instances} \\
    \midrule
      \makecell{Module \\ Entry Points} & Decorators & @ProblemDefinition @MetricDefiniton \ldots \\
    \midrule
      \makecell{Type \\ Descriptors} & Classes & \makecell{\textbf{class} Tensor \\ \textbf{class} Scalar \ldots } \\
    \midrule
      \makecell{Concepts \\ and Primitives} & \makecell{Well-known \\ Global Objects} & \makecell{ \textbf{Train}.Classify \\ \textbf{Model}.Predict \\ \textbf{Test}.Compare \ldots } \\
    \midrule
      \makecell{AI Models} & Declarative Methods & \makecell{Pipeline \\ Linear \\ Relu \\ Softmax \ldots } \\
    \bottomrule
  \end{tabular}
  \end{center}
  \caption{Examples of SAIL Language Constructs.}\label{tab:language}
\end{table}

The module script, rather than directly executed in a Python interpreter, is first sent to the SAIL parser. The SAIL parser substitutes the actual execution logic with computation nodes and connects the nodes with computational dependencies to construct the computation graph, similar to the tape-recording technique in automatic differentiation frameworks~\cite{paszke_automatic_nodate}. The parser then analyzes the computation graph and synthesizes actual benchmarking code. 
The eDSL provides its own type system with both tensors and symbolic equations as first-class citizens, and helper functions to help connect different modules. In fact, with proper type inference, there is even no need to explicitly declare the input/output types of a module.

\begin{figure}[h!]
  \includegraphics[width=0.48\textwidth]{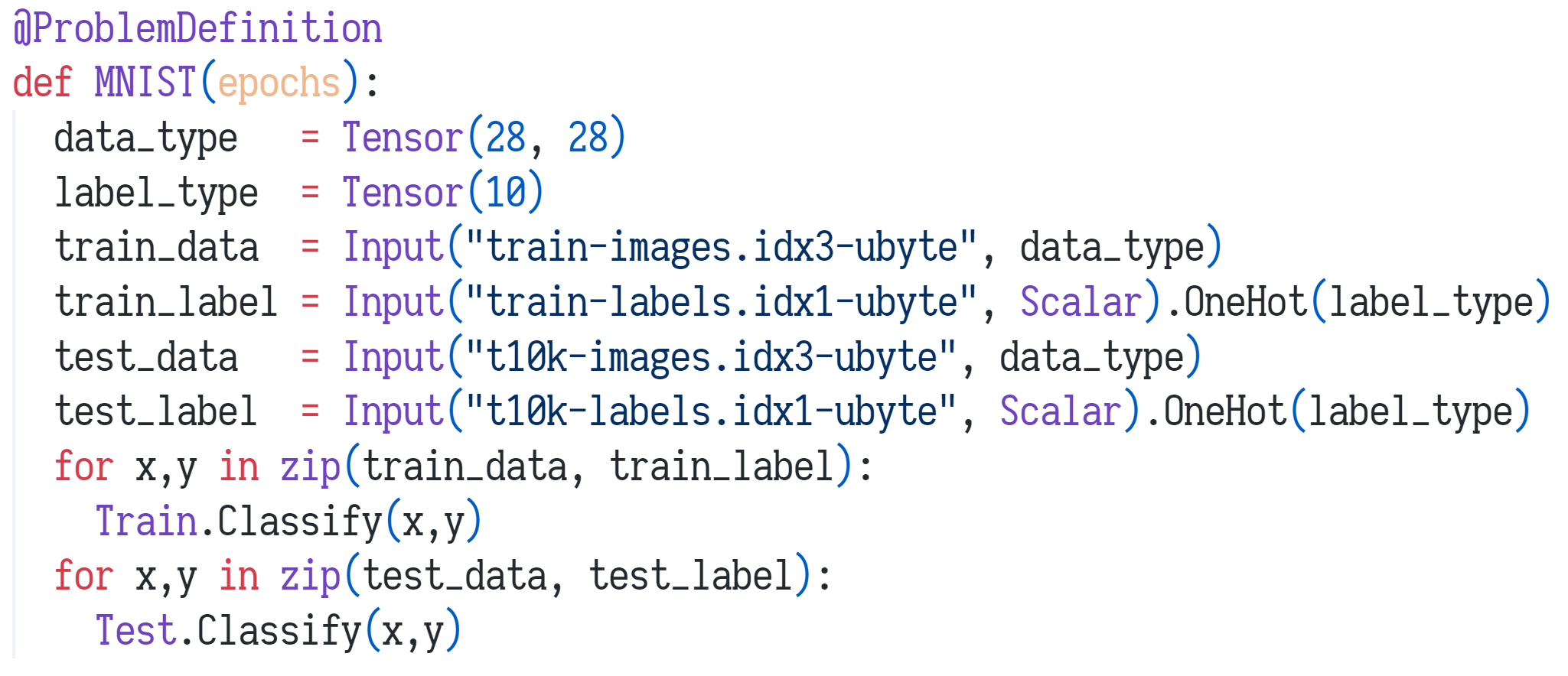}
  \caption{MNIST Problem Definition.}\label{fig:mnist}
\end{figure}

The flexibility of a scripting language also simplifies module definitions, for example, figure~\ref{fig:mnist} illustrates a ``hello world'' problem definition module --- the MNIST~\cite{lecun_gradient-based_1998} image classification problem. This is a typical ``defined by cases'' problem as we illustrated in Section~\ref{sec:problem}. The definition of this problem reads from four input files, joins them into pairs, and declares the data points and associated classification tasks into \textbf{Train} and \textbf{Test} collections. Note that the existence of both train and test collections is not necessary for some kinds of problems --- for example, a PDE ``problem class'' definition may define a few equations in the test collection and expect a solver to accomplish the task without training or hints.

\begin{figure}[h!]
  \includegraphics[width=0.4\textwidth]{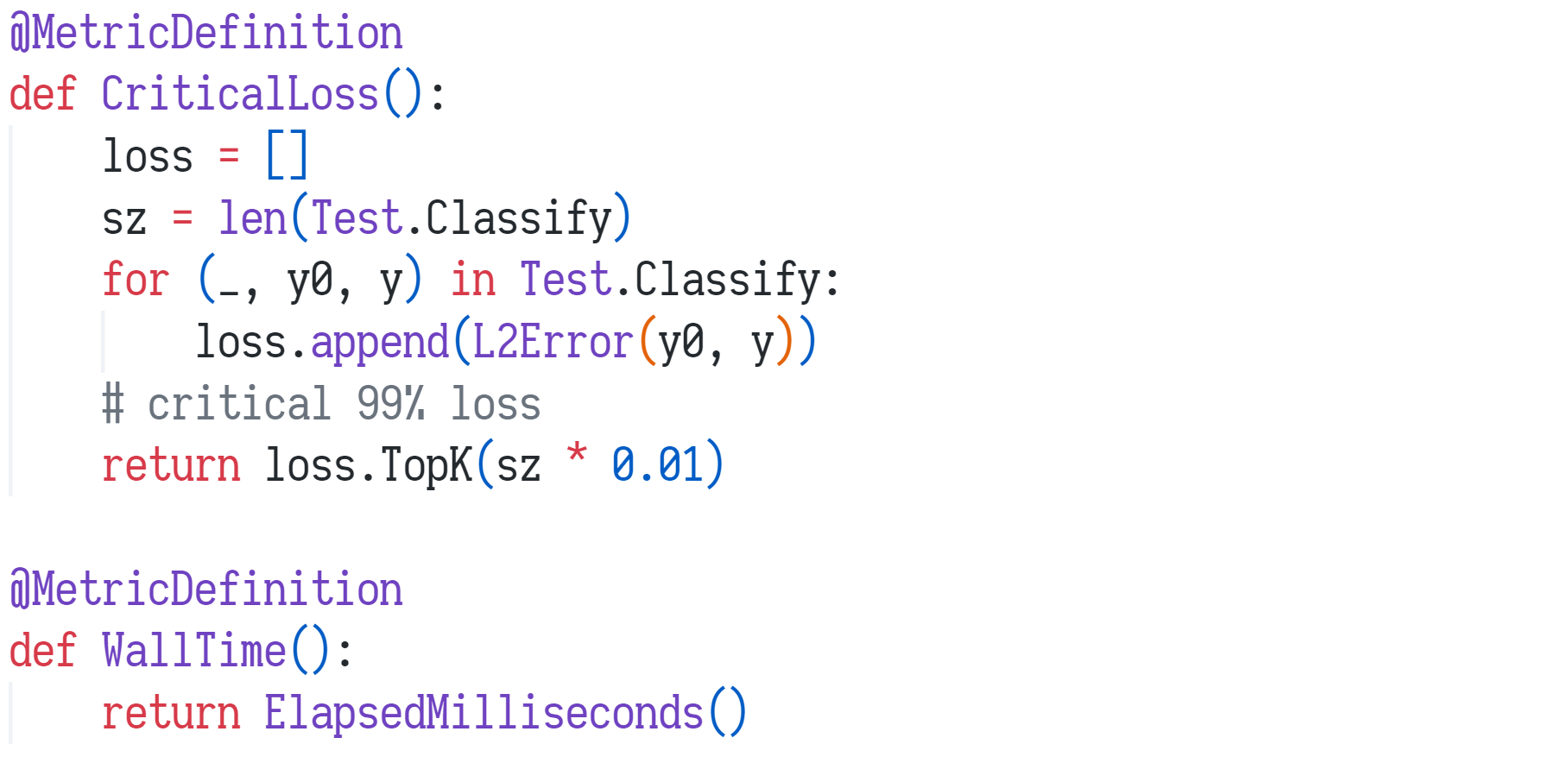}
  \caption{Custom Evaluation Metric Definition.}\label{fig:metric}
\end{figure}

Note that the problem definition of MNIST resembles a machine learning training loop --- but not entirely. The key point is that it only defines the problem, and does not try to solve or evaluate the results. This allows us to plug different evaluation metrics into the workflow. For example, the Machine Learning community traditionally focuses on the average performance over the whole dataset, while in a production critical environment, one may prefer to evaluate 99\% percentile precision, or a hard fail condition, as shown in figure~\ref{fig:metric}. Also shown in the code is a simple timer metric, and a task can be evaluated with multiple metrics. For example, the two in the code will combine into a work--critical loss 2D graph. For iterative tasks, a metric will also be evaluated iteratively, and a module can choose to keep states across multiple iterations, memorize the data points or obtain the average, etc.

Even for the same task, different research communities have different interests in performance evaluation. For example, scientific research groups focus on the quality of the end result, while computer system researchers focus on system performance metrics, such as throughput and latency~\cite{thiyagalingam_scientific_2021}. This is why we further split the ranking module from problems and metrics. A ranking module can reference multiple metrics and aggregate them to obtain a total order, or implement a comparison between two instances to obtain partial order.


\begin{figure}[h!]
  \includegraphics[width=0.48\textwidth]{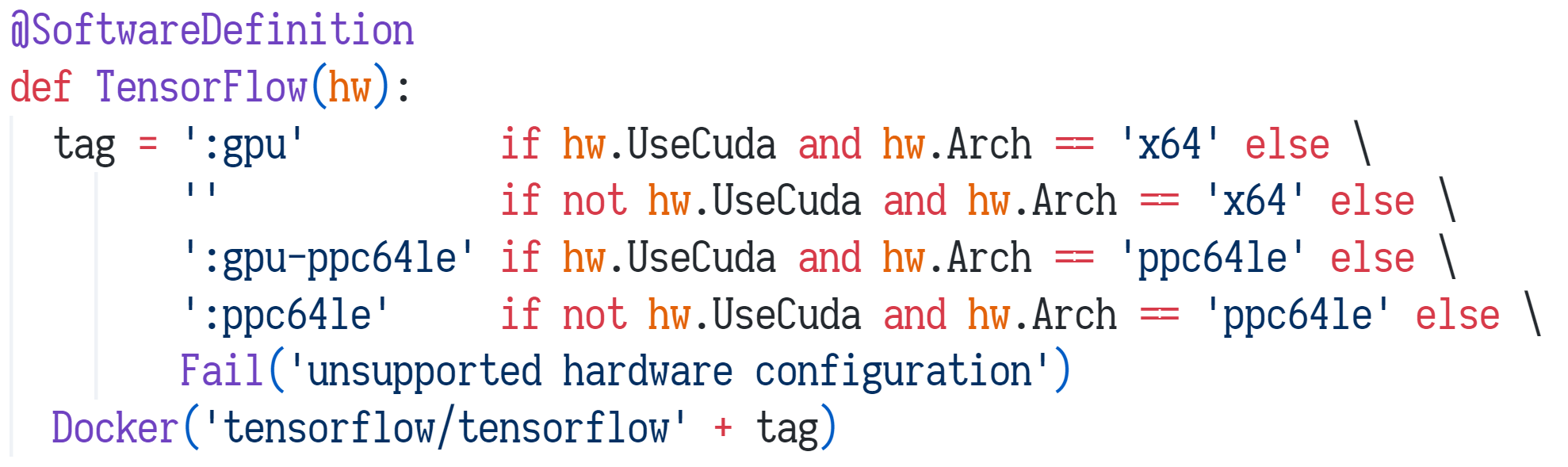}
  \caption{TensorFlow Software Configuration.}\label{fig:tf}
\end{figure}

\begin{figure}[h!]
  \includegraphics[width=0.42\textwidth]{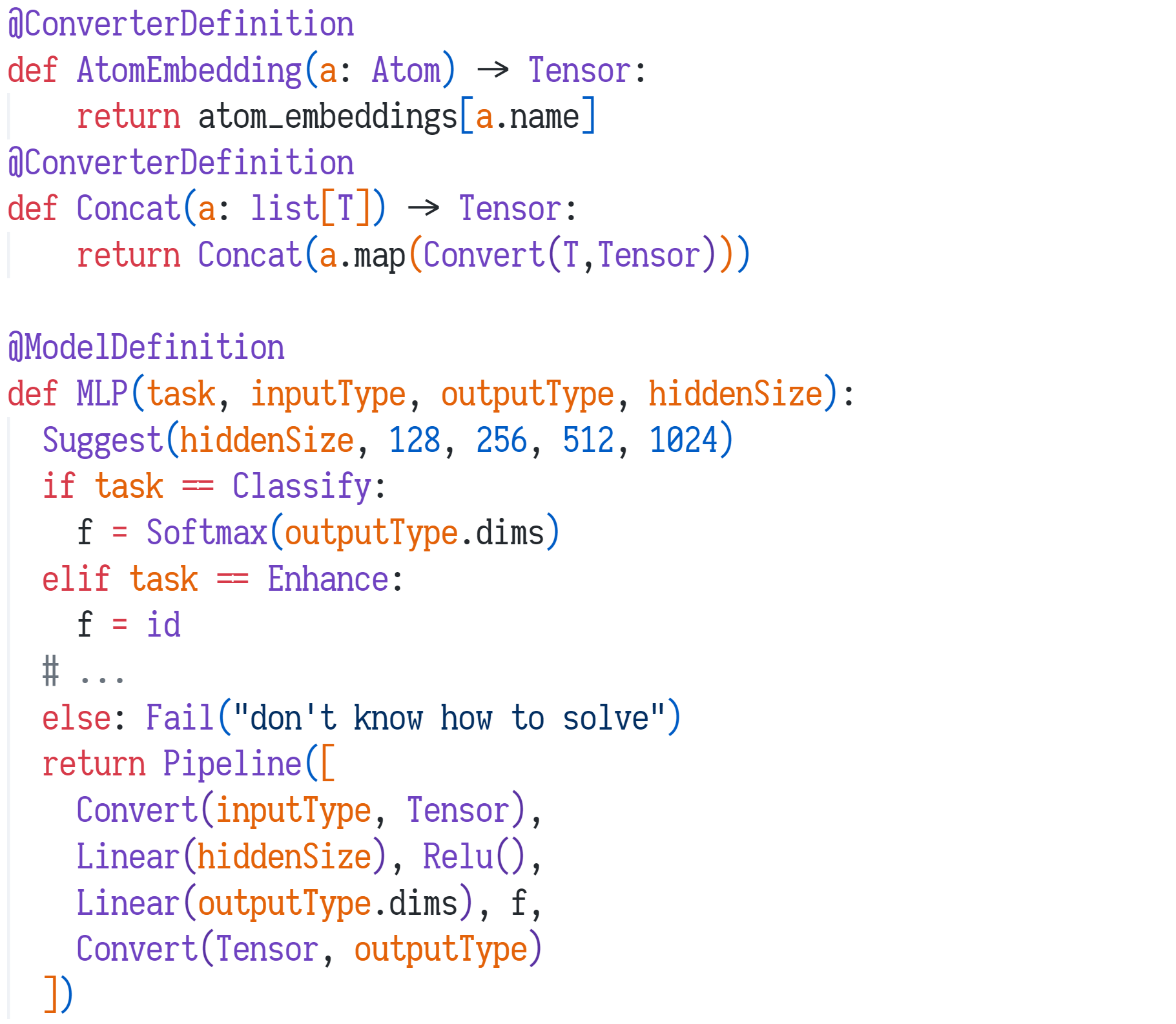}
  \caption{AI Model Definition.}\label{fig:model}
\end{figure}

Another advantage of this approach is that the module definition can take input parameters and programmatically generate configurations. For example, in figure~\ref{fig:tf} we define how to pick the correct docker image tag for TensorFlow based on the hardware configuration, which is hard to model with a markup language. This also allows us to define generic AI modules that adapt to different input sizes and types and suggest hyperparameter values. Figure~\ref{fig:model} shows the definition of a simple neural network, which not only defines the computation graph, but also the intended tasks, input/output type conversion, and layer width suggestions, so that the planner can grid search this hyperparameter. Also shown in the code snippet are two type converters, when combined, can automatically convert the input of an atom sequence into a single concatenated tensor.

\subsection{Automatic Benchmarking Task Discovery}

As discussed before, the module definitions are not used for the actual execution of the benchmarking tasks. Rather, they are metaprogramming constructs that can be seen as a ``dry-run'' for the actual benchmarking. The system scans all python files and uses reflection to identify module entry points, and create records for them in the module repository. The system then enumerates all the modules from the repository and constructs candidate test fixtures, which are tuples of different kinds of modules. For each candidate tuple, the system executes the modules in it, providing input parameters, and extracting information such as the problems a model can solve, the research field of a problem, the suggested hyperparameters, and compatible metrics for a kind of task and so on. The execution order is determined by the type of modules and the implied dependencies -- problem definitions execute first because they generally do not depend on other modules, and populate the metadata required to associate metrics and ranking. During execution, the system maintains the context for the current test fixture and accumulates the metadata from already executed modules in the candidate tuple, and later modules can either be filtered by metadata matching (for example, by matching data types) or actively reject the context. This is demonstrated in earlier examples, where a module can use the DSL primitive \textbf{Fail} to indicate that it does not know how to solve the problem, or the hardware does not support the current software configuration. Additionally, the system builds a graph where the nodes are data types and edges are converters, and employs breadth-first search to also allow type converter composition so that multiple converters can work together to relax type constraints and improve module compatibility. This process is akin to the inner-join operation in relational databases, and the system builds complete tuples of the modules as test scenarios.
Apart from automatic discovery, a module can also explicitly declare relationships with other modules so as to narrow down the search space. The logic is presented in algorithm~\ref{fig:join}.

\begin{algorithm}
  \KwIn{eDSL source files $[\mathit{src}]$}
  \KwOut{Test scenarios $[\mathit{test}]$}
  $\mathit{repo} \gets \phi$\;
  \tcp{1. Module discovery}
  \ForEach{$s : \mathit{src}$}{
    $\mathit{ast} \gets \mathrm{parse}(s)$\;
    \ForEach{$m : \mathrm{methods}(\mathit{ast})$}{
      \If{$\mathrm{decorated}(m, \mathrm{ProblemDefinition})$}{
        $\mathit{repo}[0].\mathrm{append}(m)$\;
      }
      \If{$\mathrm{decorated}(m, \mathrm{MetricDefinition})$}{
        $\mathit{repo}[1].\mathrm{append}(m)$\;
      }
      \tcp{Scan for other modules\ldots}
    }
  }
  $\mathit{ctx} \gets \phi$; $\mathit{test} \gets \phi$; $\mathit{iters} \gets \mathrm{iterators}(\mathit{repo})$\;
  $\mathit{i} \gets 0$\;
  \tcp{2. Task discovery}
  \While{$i \geq 0$}{
    \If{$i = N$}{
      $\mathit{test}.\mathrm{append}(\mathit{ctx})$\;
      $\mathit{ctx}.\mathrm{pop}()$\;
      $i\gets i-1$\;
    }
    \ElseIf{$\mathrm{next}(\mathit{iters}[i])$}{
      $m \gets \mathrm{get}(\mathit{iters}[i])$\;
      $\mathit{metadata} \gets \mathrm{execute}(m, \mathit{ctx})$\;
      \If{$\mathrm{not}\ \mathrm{Failed} (\mathit{metadata})$}{
        $\mathit{ctx}.\mathrm{push}(\mathit{metadata})$\;
        $i \gets i+1$\;
      }
    }\Else{
      $\mathit{ctx}.\mathrm{pop}()$\;
      $\mathrm{rewind}(\mathit{iters}[i])$\;
      $i \gets i-1$\;
    }
  }
  \Return{$\mathit{test}$}\
  \caption{Benchmarking Task Discovery.}\label{fig:join}
\end{algorithm}

\subsection{Experiment Orchestration}

When the planner is done generating benchmarking configuration tuples, it is necessary to prune unnecessary entries and make a schedule for the rest. There are multiple invariances in the benchmarking tasks to help pruning. For example, given the same AI model, software/hardware configuration, and similar problem size (of different problems), the throughput (in terms of FLOPS) can be comparable. Likewise, the precision evaluation should not be heavily impacted for the same model and problem on different software/hardware configurations. The executor should only pick significant tuples to maximize the diversity in measurements for all the metric dimensions, including model performance, scalability, generalization, and so on. Once the pruning is done, the scheduling problem concerns how to estimate the costs of each tuple, and efficiently pack them onto the hardware-task timeline.

\section{Case Study}\label{sec:casestudy}

Now we discuss the details of a particular use case, Molecular Dynamics (MD). Given the initial states of the atoms (position and velocity vectors), the problem asks for a prediction of the movement of the atoms. In practice, the problem is decomposed into the problem of force prediction (Molecular Mechanics), and the integrated force over time to compute new states. In particular, force prediction is achieved in multiple ways developed by the Molecular Dynamics research community. ``Classical MD'' employs empirical models to compute pairwise forces between atoms, and first-principle methods  (AIMD) employ quantum mechanic methods as DFT~\cite{haunschild_comprehensive_2019} and CCSD(t)~\cite{kummel_biography_2003} to first predict the potential energy of the system, and then obtain the forces by computing the partial derivatives of the energy over atom positions.

\begin{figure}[h!]
  \includegraphics[width=0.525\textwidth]{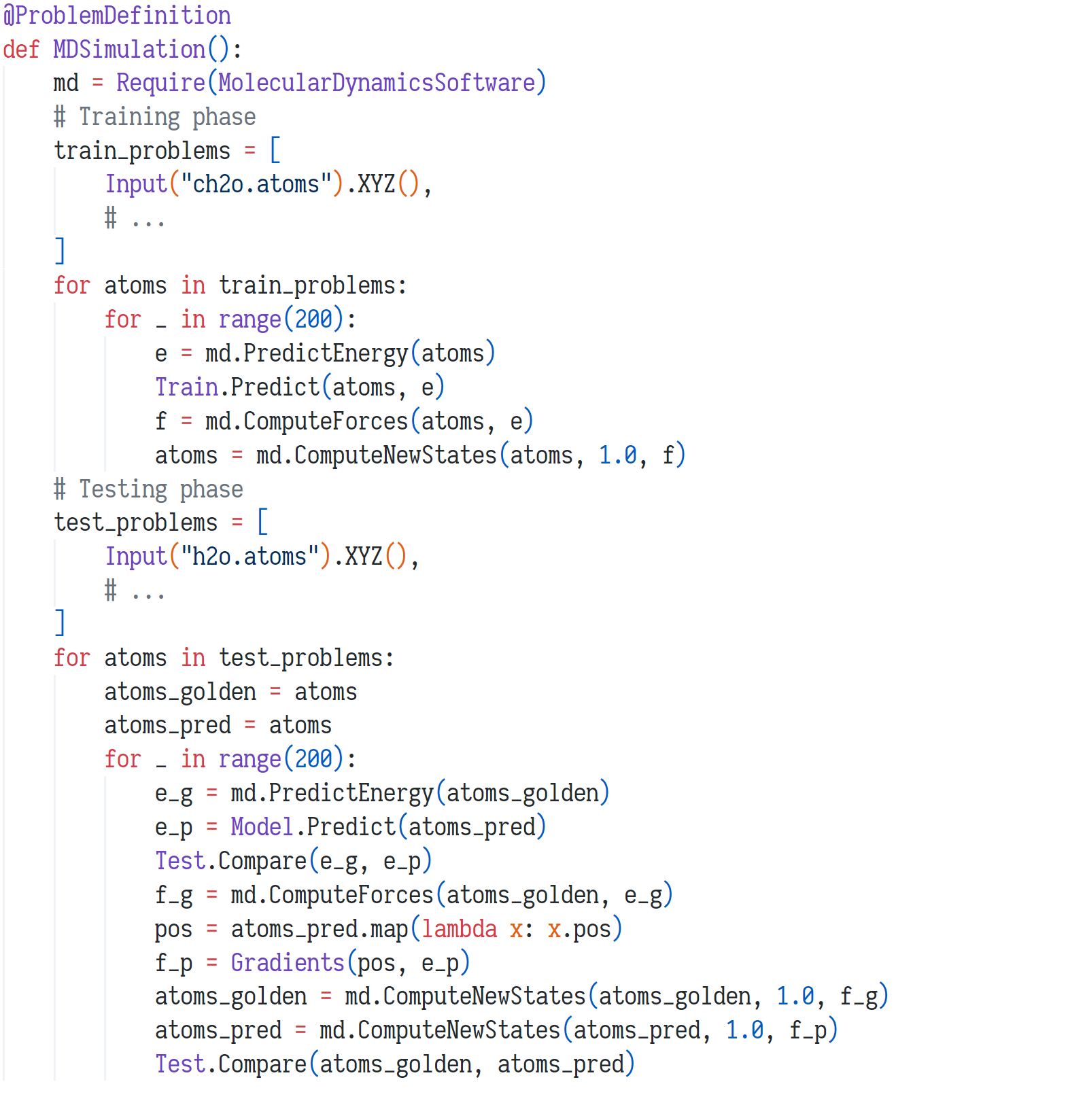}
  \caption{Molecular Dynamics Problem Definition}\label{fig:md}
\end{figure}

The problem definition module is shown in figure~\ref{fig:md}. It consists of two phases. First, an AI model is trained to predict the potential energy of a system, guided by a Molecular Dynamics software package, such as ORCA or Gaussian. Then, the performance of the model is evaluated on a different set of atom configurations. Unlike traditional AI benchmarks that merely evaluate the output of a model, here we provide multiple fixtures, including both the energy prediction, and the position and velocity updates computed from the prediction. It is the flexibility of problem scripting that gives us the ability to model additional fixtures other than the energy, which can be extended to benchmark fields other than Molecular Dynamics, for example, Raman Spectroscopy. This is a typical ``defined by setting'' problem as we illustrated in Section~\ref{sec:problem}, because although it reads multiple data points from input files, each data point is not fed into the AI model, but rather into a simulation software to compute data points for the AI model.

\begin{figure}[h!]
  \includegraphics[width=0.525\textwidth]{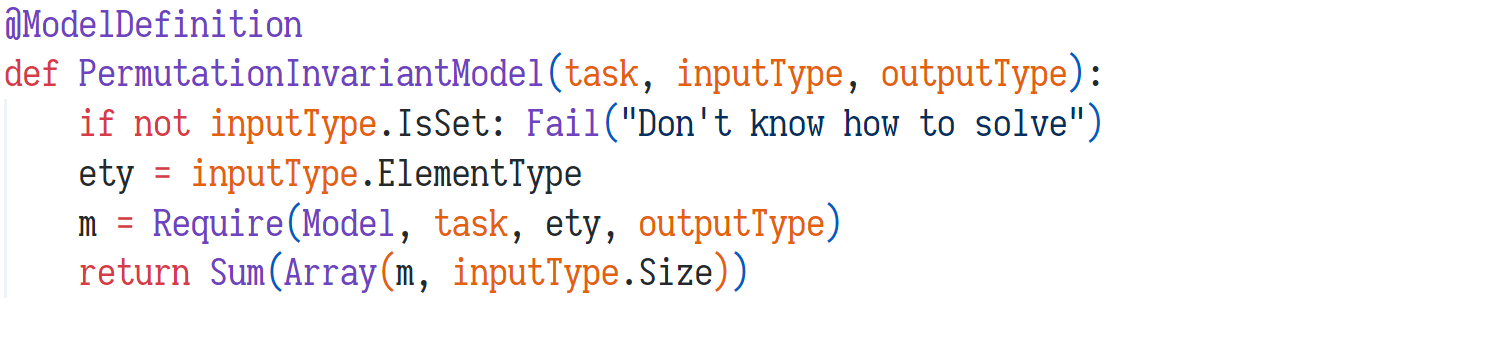}
  \caption{Permutation Invariant Model Definition}\label{fig:perm}
\end{figure}

There are multiple ways to specify an AI model for this problem -- namely, given a set of atoms (atom types, positions, and velocities), predict a single scalar energy value. One way is to implement an end-to-end energy prediction model~\cite{han_deep_2018}~\cite{unke_physnet_2019}. The other way aims to capture the essence of the end-to-end solutions and let the system synthesize the whole model. One key insight of the aforementioned energy prediction models is that the atom configuration is permutation invariant, which means that the input should be modeled as a set of atoms, not a list. Therefore, our goal here is to enable the system to compose an AI model to honor this property and take advantage of existing building blocks. A possible solution is shown in figure~\ref{fig:perm}, where the input is typechecked to be a list, and the module requires a submodule that can complete the specified task (prediction in the Molecular Dynamics context) to map the element type to the output type. The element-wise results are then summed to combine a permutation invariant output. This way, the system is able to pick up the modules we defined earlier, such as the atom embedding converter, and the MLP model for conducting element-wise prediction.

\begin{figure}[h!]
  \includegraphics[width=0.46\textwidth]{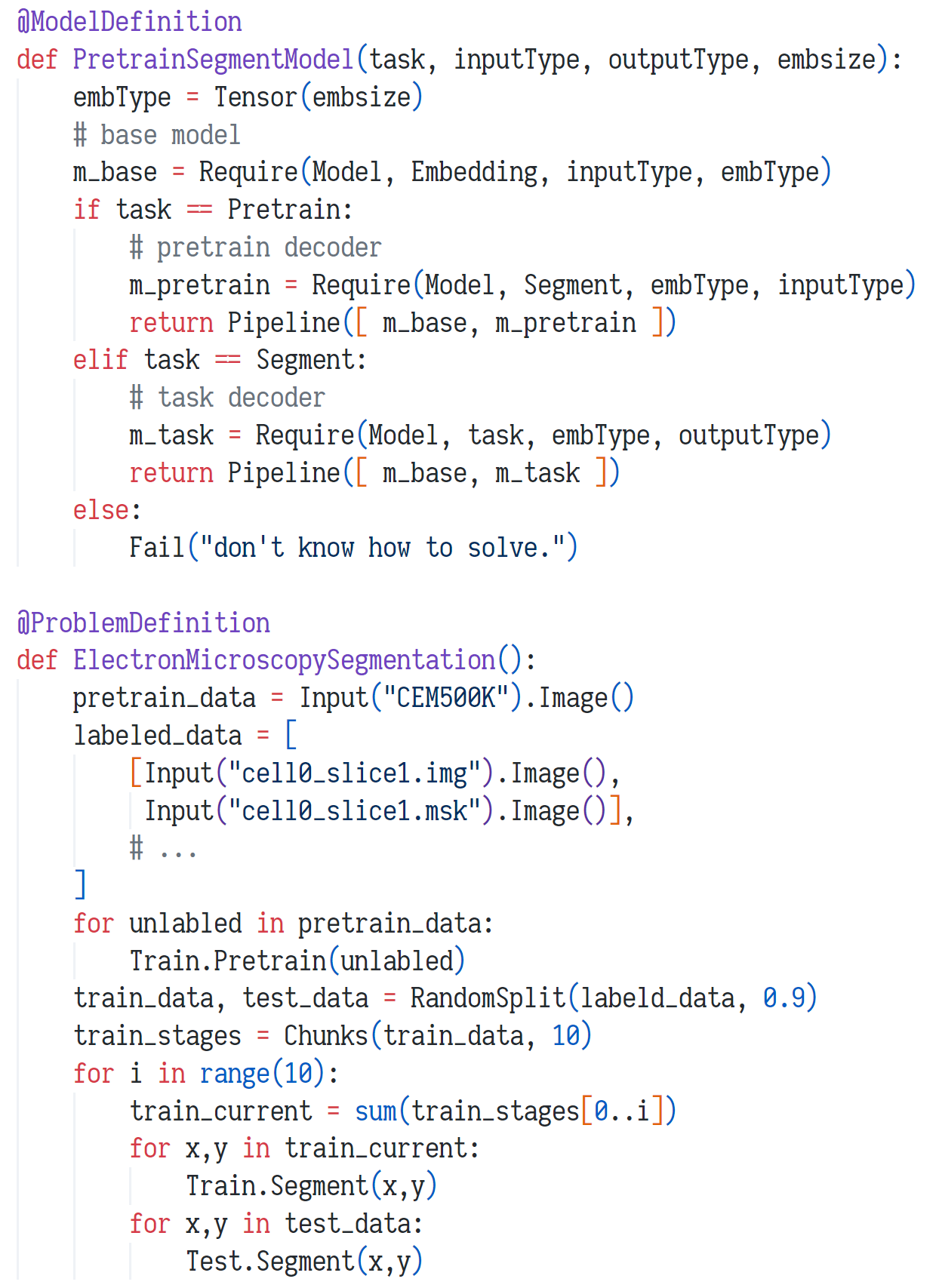}
  \caption{Electron Microscopy Image Segmentation.}\label{fig:pretrain}
\end{figure}

Now we discuss another benchmarking scenario, deep-learning-based electron microscopy image segmentation, which is becoming a popular topic in Biological Chemistry~\cite{gomez-de-mariscal_deep-learning-based_2019}\cite{von_chamier_democratising_2021}\cite{ede_deep_2021}. One of the main challenges in this topic is the scarce of training data, due to complex and costly data acquisition process. Given limited data, supervised deep-learning methods require heavy human intervention and may fail to generalize to unseen data~\cite{plaza_analyzing_2018}\cite{lichtman_big_2014}. One way to circumvent the data problem is to introduce semi-supervised deep-learning techniques, such as pre-training with high volume unlabeled data~\cite{conrad_cem500k_2021}. To support pre-training in the benchmarking system means that a model under evaluation should be able to carry a part of its internal states (weights) from one task to another, and adjust its computation graph accordingly. The problem definition should also evaluate the performance of the model given different amounts of training data, to test its sample efficiency. The code for this scenario is shown in figure~\ref{fig:pretrain}.

\subsection{Comparison to Other Benchmarking Systems}

As mentioned above, previous systems focus on a fixed set of test scenarios~\cite{zheng_aibench_2019}\cite{mattson_mlperf_nodate}\cite{chang_mlharness_2021}. Additionally, the lack of declarative modules means that it is hard to share data between the benchmarking suite and external scientific computing software packages, which is crucial in scientific AI benchmarking. For example, the \textbf{Gradient} primitive in \ProjName{} allows a training pipeline to extract gradients from an external package, which is usually not exposed programmatically. The differences are shown in table~\ref{tab:compare}.

\begin{table}
    \begin{tabular}{p{0.09\textwidth} p{0.1\textwidth} p{0.1\textwidth} p{0.11\textwidth}}
    \toprule
      & \textbf{\ProjName{}} & \textbf{MLPerf} & \textbf{MLHarness} \\
    \midrule
      \textbf{Focus} & Different scientific tasks/criterion & Accuracy, system throughput & Scalability, MLCommon coverage \\
    \midrule
      \textbf{Modules} & Declarative & Hard-coded & Markup \\
    \midrule
      \makecell{\textbf{Test}\\\textbf{Scenarios}} & Automatic Discovery & Fixed & Fixed \\
    \bottomrule
  \end{tabular}
  \caption{Comparison to Other Benchmarking Systems.}\label{tab:compare}
\end{table}
\section{Discussion}

We have elaborated on the methodology and the overview of the system design, yet we look forward to further development in the components. Brute-force enumeration of all possible test hyperparameters may not be feasible and while pruning can mechanically improve the situation, it is desirable that a particular problem module can suggest parameters suitable for a research field. More design work could be done to address model development and debugging needs, for example, to allow model validation in addition to training and testing. Python-based eDSL has its limitations, mostly due to the syntactic constraints of the language. To represent the modules more naturally, a programming language more geared toward scientific computing can be investigated~\cite{innes_differentiable_2019}.

Currently, \ProjName{} targets tractable scientific tasks, which are mechanical procedures that can be computed and measured. It is challenging to extend it to more creative scientific research activities because it would require the system to formally model the scientific concepts, and gain a deeper understanding of research topics, motivations, methodologies, and goals, and how various concepts interact with each other. Also, automated benchmarks require well-defined metrics, while open-ended scientific research ideas, in general, are hard to quantify.

Apart from type-based model composition, automatic AI model synthesizing given a particular problem definition is also a promising direction, given the advancement in AI-based code generation~\cite{lu_codexglue_2021}~\cite{peng_how_2021}.

\section{Conclusion}

We have presented our definition of scientific AI benchmarking, which is an ensemble of scientific task definition, AI benchmarking, and system performance benchmarking. We have then presented our methodology for scientific AI benchmarking, with the key idea of decoupling and modularizing various components, automatically benchmarking sensible combinations. We have proposed a system design where the various modules are implemented with a domain-specific language for scientific AI computing. We have demonstrated that this design is flexible enough to support benchmarking different types of scientific tasks, defining AI models, deriving multiple metrics, combining metrics into ranking criteria, and configuring required hardware/software.

\bibliographystyle{elsarticle-num}
\bibliography{SAIBench-refs}
\end{document}